\documentclass[journal]{IEEEtran}

\usepackage{amsmath,amsfonts}
\usepackage[caption=false,font=normalsize,labelfont=sf,textfont=sf]{subfig}
\usepackage{graphicx}
\usepackage{cite}
\usepackage{multirow}
\usepackage{colortbl}
\usepackage[table]{xcolor}
\usepackage{diagbox}
\usepackage{hyperref}
\usepackage[capitalize,noabbrev]{cleveref}
\begin{document}

\title{Efficient Parallel Genetic Algorithm for Perturbed Substructure Optimization in Complex Network}

%
%

\author{\IEEEauthorblockN{Shanqing~Yu,~Meng~Zhou,~Jintao~Zhou,~Minghao Zhao,~Yidan~Song,~Yao Lu,~Zeyu~Wang,~Qi~Xuan,~\IEEEmembership{Senior~Member,~IEEE}}
\IEEEcompsocitemizethanks{
\IEEEcompsocthanksitem This work was supported in part by the Key R\&D Program of Zhejiang under Grant 2022C01018 and 2024C01025, by the National Natural Science Foundation of China under Grant 62103374 and U21B2001, by Key Technology Research and Development Program Project of Hangzhou under Grant 2024SZD1A23, and by the Zhejiang Provincial Natural Science Foundation of China under Grant ZCLY24F0302.
\IEEEcompsocthanksitem Shanqing Yu, Meng Zhou, Jintao Zhou, Yidan Song, Yao Lu, Zeyu Wang, Qi Xuan are with the Institute of Cyberspace Security, College of Information Engineering, Zhejiang University of Technology, Hangzhou, China; Binjiang Institute of Artificial Intelligence, Zhejiang University of Technology, 310056, Hangzhou, China (E-mail: yushanqing@zjut.edu.cn; nakilea3315@gmail.com; Zhou\_974@outlook.com; 3292551392@qq.com; yaolu.zjut@gmail.com; Vencent\_Wang@outlook.com; xuanqi@zjut.edu.cn).
\IEEEcompsocthanksitem Minghao Zhao is with the Binjiang Institute of Artificial Intelligence, Zhejiang University of Technology, 310056, Hangzhou, China (E-mail: yzbyzmh1314@163.com).
\IEEEcompsocthanksitem Corresponding author: Zeyu Wang.
}
}


\maketitle
\begin{abstract}

Evolutionary computing, particularly genetic algorithm (GA), is a combinatorial optimization method inspired by natural selection and the transmission of genetic information, which is widely used to identify optimal solutions to complex problems through simulated programming and iteration. Due to its strong adaptability, flexibility, and robustness, GA has shown significant performance and potentiality on perturbed substructure optimization (PSSO), an important graph mining problem that achieves its goals by modifying network structures. However, the efficiency and practicality of GA-based PSSO face enormous challenges due to the complexity and diversity of application scenarios. While some research has explored acceleration frameworks in evolutionary computing, their performance on PSSO remains limited due to a lack of scenario generalizability. Based on these, this paper is the first to present the GA-based PSSO Acceleration framework (GAPA), which simplifies the GA development process and supports distributed acceleration. Specifically, it reconstructs the genetic operation and designs a development framework for efficient parallel acceleration. Meanwhile, GAPA includes an extensible library that optimizes and accelerates 10 PSSO algorithms, covering 4 crucial tasks for graph mining. Comprehensive experiments on 18 datasets across 4 tasks and 10 algorithms effectively demonstrate the superiority of GAPA, achieving an average of 4x the acceleration of Evox. The repository is in \url{https://github.com/NetAlsGroup/GAPA}.

\end{abstract}

\begin{IEEEkeywords}
Genetic Algorithm, GPU Acceleration, Perturbed Substructure Optimization, Graph Mining.
\end{IEEEkeywords}

\section{Introduction}
\IEEEPARstart{E}{volutionary} computation is an algorithm inspired by the natural evolutionary process, that can explore the optimal solution of complex combinatorial optimization problems by simulating the biological evolutionary process, with high adaptability and flexibility\cite{pena2000evolutionary, malik2021metaheuristic}. Among them, the evolutionary algorithms represented by Genetic Algorithm(GA) prefer the natural selection mechanism and the law of genetic information transfer in biological evolution, which is widely used in neural evolution\cite{stanley2019designing, liu2023egnn, liu2024constrained}, natural language processing\cite{onan2023gtr}, graph mining\cite{9172881, tde, lpa-eda}, etc. Particularly, it has achieved excellent performance on the perturbed substructure optimization (PSSO) problem, which is a typical combinatorial optimization problem in graph mining.

PSSO is a task that aims to achieve specific goals by modifying the network structures with a perturbation budget\cite{morone2015influence}, covering various applications such as critical node detection\cite{liu2019framework, tde, cutoff, zhang2023interactive, zhang2020multi, liu2024vital}, community detection\cite{9172881, 8714065, cgn, chen2020multiscale, zhao2023self, huang2021higher}, link prediction\cite{lpa-eda}, node classification\cite{gani, dai2018adversarial}, etc. Essentially, PSSO is a problem that searches for optimal perturbed substructures from huge topological combinations, and GA has gained wide attention for its powerful global search capability. For example, Chen et al.\cite{8714065} used GA to implement a community detection attack. Yu et al.\cite{9172881} proposed an Euclidean distance attack method using a genetic algorithm to scramble edges and thus turn off downstream data mining. Fang et al.\cite{dai2018adversarial} injected fake nodes through GA to affect the ability of node classification models. However, the increasing complexity of application scenarios presents significant challenges for GA-based PSSO in terms of efficiency and computational cost, putting higher demands on the efficiency and generalizability of GA as well as further optimization and acceleration. Therefore, acceleration frameworks for GA have received widespread attention. GPU hardware offers powerful parallel computing capabilities, making it a vital tool for accelerating deep learning model training tasks\cite{buber2018performance}. Many researchers have explored the use of GPUs to accelerate evolution algorithms with promising results\cite{li2007efficient, cheng2020parallel, evojax, evosax, evotorch, evox}. Among them, Evotorch\cite{evotorch} provides an acceleration framework based on PyTorch for evolutionary algorithms. However, it lacks architectural design flexibility and a unified programming model, making it challenging to implement general-purpose evolutionary algorithms efficiently\cite{evox}. Evox\cite{evox} introduces a scalable and user-friendly acceleration framework for evolutionary computation based on JAX to address these limitations. While JAX offers powerful capabilities, it requires precompilation and transformation of functions, such as fitness functions. JAX also suggests using pure functions that neither read nor modify external states. Meanwhile, it discourages using iterators in functions, aligning with its functional programming paradigm. However, in PSSO tasks, a graph is often subject to disturbances, necessitating GA to read and modify state variables and perform iterative computations within fitness or auxiliary functions. Additionally, libraries like Evotorch and Evox excel in general-purpose evolutionary computation but are not explicitly designed for GA-based PSSO problems. The genetic operation in GA, particularly for PSSO tasks, differ significantly from traditional evolutionary algorithms. For example, Evox does not support complex mutation operations. Porting such operations to the JAX framework demands a deep understanding of JAX, which poses a significant challenge for users. As a result, these frameworks may not be applicable or rapidly deployable when solving domain-specific problems like PSSO.

To address these challenges, we propose GAPA, a GA acceleration framework based on PyTorch customized for the PSSO problem. For that, we reconstruct the genetic operation and design general acceleration modes. Specifically, on the genetic operation side, we reconstruct and simplify the genetic operation. Especially for the fitness function in the genetic operation, considering its diversity, we propose a strategy for designing a fitness function with the goal of efficient parallel calculation. On the acceleration mode side, We propose four computation modes based on different levels of parallelism to satisfy the demands of various algorithms and common applications. Finally, we organize the above work into a generic functional programming framework, covering 4 PSSO tasks and 10 GA-based PSSO algorithms, which supports fast and easy deployment of GA-based PSSO algorithms. To demonstrate the superiority of GAPA in parallel acceleration, we utilize it to accelerate 10 GA algorithms across these four tasks with 12 datasets. The main contributions of this work are as follows:   

\begin{itemize}
    \item We are the first to propose and implement a customized GA-based PSSO acceleration framework, GAPA, which implements parallel acceleration for GA from three perspectives: genetic operation, fitness function design, and acceleration mode design. Meanwhile, GAPA supports simple and functional programming for PSSO algorithms.
    \item Based on the GAPA, we have built and publicized a GA acceleration library for the PSSO problem for the first time. The library covers 4 important graph mining tasks, contains 10 PSSO algorithms, and is highly extensible.
    \item Extensive experiments are conducted on multiple real-world datasets to evaluate the effectiveness of the proposed method. The results show that, while retaining high-quality solutions, the GAPA framework significantly accelerates the PSSO problem under the genetic algorithm. Specifically, compared to the state-of-the-art baseline Evox, GAPA achieves nearly 4$\times$ further acceleration.
\end{itemize}


\section{Related Work}
\label{Related Work}


\subsection{GA-based PSSO}

PSSO is a task that modifies the network structure to achieve specific objectives, which plays a crucial role in graph mining. Fundamentally, PSSO aims to identify the optimal perturbed substructure from the huge topology combination space, with GA receiving significant attention for its robust global search capabilities. In this context, we focus on GA-based PSSO problems in community detection attack (CDA), critical node detection (CND), node classification attack (NCA), and link prediction attack (LPA). Specifically, \textbf{CDA} aims to modify the network structure to confuse community detection algorithms, such as \cite{8714065, cgn, 9172881} utilize the community modularity, normalized mutual information, and network topology to explore perturbation structures and apply them in social networks to preserve privacy. \textbf{CND} focuses on identifying critical nodes in networks\cite{gao2012networks, 2018The, lokhov2014inferring}, with recent approaches utilizing network connectivity\cite{tde} and pruning strategies\cite{cutoff}, widely applied in traffic and power networks, etc. \textbf{NCA} aims to confuse classifier or intentionally perturb the classification prediction by modifying the network structure. Recent research has employed prediction confidence\cite{dai2018adversarial} and fake node injection\cite{gani} in a black-box setting, achieving precise perturbations in citation networks. \textbf{LPA} manipulates link prediction results for attacks or privacy protection in complex networks, with recent research\cite{lpa-eda} seeking network-sensitive edges to facilitate efficient privacy protection in social networks.

Overall, GA-based PSSO research holds considerable practical significance. However, as PSSO tasks grow more diverse and intricate, coupled with the increasing scale of data across various application domains, these methods encounter substantial challenges in sustaining both algorithmic efficiency and real-world applicability.

\begin{figure*}[htp]
\centering
\includegraphics[width=\textwidth]{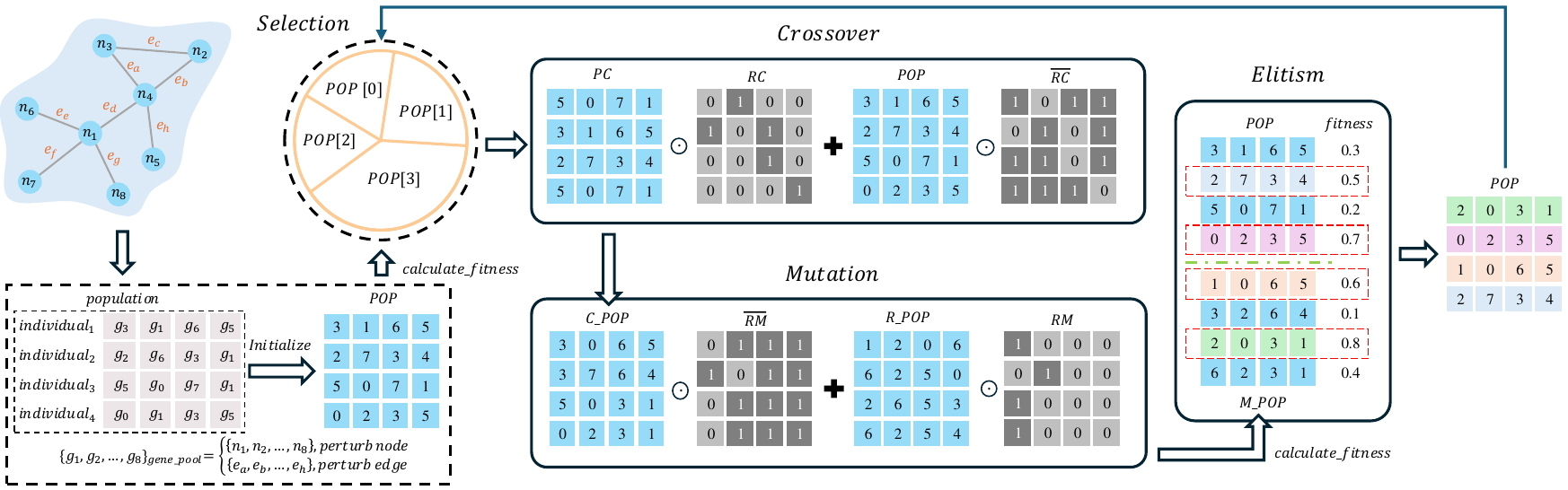}\\\
\caption{Acceleration framework of GAPA in genetic operation. GAPA is an acceleration framework designed to optimize genetic operation, including population initialization, crossover, mutation, fitness calculation, and elitism. Specifically, this figure sets the elite size equal to the initial population size and takes a network of eight nodes as an example to illustrate the optimization strategy of GAPA. In this context, the gene pool can be composed of network edges or nodes according to the type of perturbation, represented by $\{ g_1, g_2, \cdots, g_8\}$, $individual_i$ represents the genome owned by the perturbation, and $population$ represents the perturbation population. The initialization operation transforms the perturbation population into a perturbation matrix $POP$, performing crossover and mutation through simple matrix operations. Finally, based on the results calculated by the fitness function, elitism is applied to retain the individuals with the best perturbation effects, which are then used in the next iteration.}
\label{fig:Main Architecture}
\end{figure*}

\subsection{Accelerating GA}

Genetic algorithms, a kind of classic evolutionary algorithms, are inherently well-suited for parallel computing due to their population-based approach, which allows candidate solutions to be evaluated independently, making them highly compatible with hardware acceleration\cite{evox}. Traditional methods for parallelizing GA on CPU architectures, such as the Master-Slave, Island, and Cellular models\cite{10.5555/645513.657737, chipperfield1994parallel}, remain widely used, but as the demand for algorithm efficiency increases, traditional methods are still limited. Pygmo \cite{pygmo} employs a generalized island model with coarse-grained parallelization, enabling large-scale parallel optimization. Furthermore, with recent advancements in hardware, GPUs have achieved remarkable success in machine learning acceleration algorithms, which provide a new perspective for including evolutionary algorithms acceleration\cite{CHENG2019514}. For instance, Li et al.\cite{li2007efficient} proposed a method for fine-grained parallel acceleration of GA. Meanwhile, inspired by the powerful computing capabilities of GPUs, several pioneering evolutionary computation acceleration libraries have emerged. EvoJAX\cite{evojax}, leverages the JAX framework to integrate evolutionary computation with neural networks, advancing scalable, hardware-accelerated neural evolution. Building on EvoJAX, evosax\cite{evosax} focuses on evolutionary strategies and offers a curated set of GPU-optimized algorithms.
Similarly, EvoTorch\cite{evotorch} capitalized on PyTorch’s strengths to develop an evolutionary computation acceleration library. However, Evox\cite{evox} argues that deep learning-centric frameworks like PyTorch may not be ideal for accelerating evolutionary computation and that previous libraries lack a unified programming and computational model. To address these issues, Evox introduced a unified workflow based on JAX, featuring a simple functional programming model that significantly reduces development complexity.

In summary, these libraries provide valuable and efficient frameworks for accelerating evolutionary computation. However, libraries like EvoTorch, which is built on PyTorch, lack architectural design flexibility and a unified programming model, making it challenging to implement general-purpose evolutionary algorithms efficiently, while the features of JAX make it unsuitable for handling the frequent disturbances of the PSSO problem. Most critically, these libraries are designed primarily for single-objective optimization, multi-objective optimization, and neuroevolution, making them unsuitable for GA-based PSSO problems.

\section{Methodology}
\label{Methodology}
Here, we elaborate on a GA acceleration framework for the PSSO problem, focusing on the genetic operation, fitness function design, and acceleration mode design, to effectively accelerate the GA-based PSSO problems.

\subsection{Problem Definition of PSSO}
\label{PSSO}
Formally, for the PSSO problem, we assume an undirected and unweighted network $\mathcal{G}=\left(\mathcal{V},\mathcal{E}\right)$, where $\mathcal{V}=\{v_1,v_2,\ldots,v_n\}$ denotes a set of $n$ nodes and $\mathcal{E}=\{e_1,e_2,\ldots,e_m\}$ denotes a set of $m$ edges. Based on this, we define the PSSO problem as follows:

\begin{equation}
\begin{array}{l}

\begin{aligned}
 & \mathit{arg}\ \mathop{\mathrm{max}}\limits_{\Delta f} \text{S}\bigg( {\Delta f(\mathcal{G},\mathit{p})} \bigg), \\
 & \begin{aligned}
   \mathit{s.t.}\; & C(p), \\
 & p\in\mathcal{P}
   \end{aligned} 
\end{aligned}

\label{eq:optpro} 
\end{array}
\end{equation}
where the objective of the PSSO problem is to find a perturbation $p$ of the original network $\mathcal{G}$ in the set $\mathcal{P}$ containing all perturbed substructures satisfying the given constraints. The perturbation change of the target network is denoted as $\triangle f\left(\mathcal{G},p\right)$, and $C\left(p\right)$ denotes the cost of the perturbation that needs to be controlled. Facing different application scenarios, the policy function $\text{S}(\cdot,\cdot)$ can be designed in different forms.

\subsection{Accelerating Genetic Operation}
\label{Architecture}

GA is an optimization algorithm based on natural selection and genetic mechanisms. Its basic operation includes five key operators: population initialization, crossover, mutation, fitness calculation, and elitism. Although genetic algorithms offer significant advantages in solving complex problems, their traditional framework often requires multiple loop updates in each iteration, leading to low computational efficiency, especially for large-scale problems with a significantly limited iteration speed. To address this problem, we proposed GAPA, which converts the traditional genetic operation into a form that facilitates batch and parallel calculation, that can significantly improve the execution efficiency of the algorithm.

\subsubsection{Population Initialize}
The population is initialized to generate a set of initial candidate solutions, and the scale of initial candidate solutions in the population is the same, so the initialized population can be easily converted into a matrix form.  Meanwhile, to facilitate subsequent genetic calculations, we convert this set into a candidate solution matrix. Specifically, the initial population is randomized, where each solution can be represented as a vector $individual_i= \left[{ind}_{i1},{ind}_{i2},\ldots,{ind}_{ik}\right]$, with $i$ represents the index of the individual in the population, and ${ind}_{ik}$ represents the $k$-th gene of the current individual. In practice, each possible gene is assigned a unique index for computation. And the initial population always has $s$ individuals and $k$ genes, represented as $POP_0 \in \mathbb{R}^{s \times k}$:

\begin{equation}
POP_0=\left[\begin{matrix}individual_1\\\vdots\\individual_s\\\end{matrix}\right]=\left[\begin{matrix}{ind}_{11}&\cdots&{ind}_{1k}\\\vdots&\ddots&\vdots\\{ind}_{s1}&\cdots&{ind}_{sk}\\\end{matrix}\right]
\end{equation}
Next, the fitness value for the current initialized population is computed using the fitness calculation operator.

\subsubsection{Crossover}


The crossover operator selects two individuals from the current population and exchanges their genes based on the crossover rate. The essence of the crossover operation is to exchange genes in individuals through gene indexes. This exchange process can be easily simplified into matrix operations. Combined with the generated matrix $POP_0$, we simplify the crossover operation into a matrix operation. Specifically, the gene exchange at the crossover position is performed based on a binary mask matrix, generated using the preset crossover rate $pc$, referred to $RC \in \mathbb{R}^{s \times k}$:

\begin{equation}
    RC=\left[\begin{matrix}{TC}_{11}&\cdots&{TC}_{1k}\\\vdots&\ddots&\vdots\\{TC}_{s1}&\cdots&{TC}_{sk}\\\end{matrix}\right]
\end{equation}

\begin{equation}
\label{eq:pc}
{TC}_{ij} =
\begin{cases}
0,&{\text{rand}}(0, 1)>pc\\
{1,}&\text{otherwise.}
\end{cases}
\end{equation}
\\
where $TC_{ij}$ indicates whether the $j$-th gene of the $i$-th individual in the population undergoes crossover, and its value is determined by \cref{eq:pc} to be 0 or 1, the crossover matrix $RC$ ensures that the overall proportion of 1 in the matrix is approximately equal to $pc$, meeting the requirements of the actual crossover rate. Furthermore, the repeated selection of two individuals for crossover transformation is restructured into the following form using a parallel optimization approach:

\begin{equation}
    PC=\text{Select}\left(POP\right)
\end{equation}

\begin{equation}
    C\_POP=PC\odot RC + POP\odot\overline{RC}
\end{equation}
where $PC$ represents the matrix obtained through roulette wheel selection $\text{Select}(\cdot)$ from the current population matrix $POP$, $C\_POP$ denotes the population after the crossover operation, while $\overline{RC}$ is the binary inversion (1 to 0 and vice versa) of the crossover matrix $RC$, the symbol $\odot$ represents element-wise matrix multiplication.

\subsubsection{Mutation}
In the mutation operation, the GA iteratively selects individuals from the population and determines whether to perform gene mutation at each site based on the mutation rate $pm$. The essence of gene mutation is to find the gene points and corresponding indexes that need to be mutated by traversal and replace the original gene points with the mutated genes. Therefore, this mutation operation can be easily converted into matrix operations. Similarly, we refer to the crossover operation to transform the mutation operation into matrix operations. Specifically, a mask matrix is generated according to the mutation rate $pm$, referred to as the mutation matrix $RM \in \mathbb{R}^{s \times k}$:

\begin{equation}
RM=\left[\begin{matrix}{TM}_{11}&\cdots&{TM}_{1k}\\\vdots&\ddots&\vdots\\{TM}_{s1}&\cdots&{TM}_{sk}\\\end{matrix}\right]
\end{equation}

\begin{equation}
\label{eq:pm}
{TM}_{ij} =
\begin{cases}
0,&{\text{rand}}(0, 1)>pm\\
{1,}&\text{otherwise.}
\end{cases}
\end{equation}
where $TM_{ij}$ indicates whether the $j$-th gene of the $i$-th individual in the population has mutated, with its value determined by \cref{eq:pm} to be either 0 or 1, the mutation matrix $RM$ ensures that the overall proportion of 1 in the matrix approximately equals $pm$, satisfying the requirements of the actual mutation rate. Similarly, the repeated selection of individuals for mutation is transformed as follows:

\begin{equation}
    M\_POP=C\_POP\odot\overline{RM}+R\_POP\odot RM
\end{equation}
where $R\_POP$ is the mutation index matrix, which will generate the corresponding mutant gene according to the index, $\overline{RM}$ is obtained by inverting each entry in the mutation matrix RM, the matrix $M\_POP$ obtained after crossover mutation represents the newly generated population matrix of the current generation.

\subsubsection{Fitness Calculation}
Considering the diversity and complexity of fitness functions, we will detail the fitness function design process in \cref{SixDST}.

\subsubsection{Elitism}
Elitism retains the dominant individuals in the population, as its essence is an operation based on sorting and comparison. To ensure the consistency of data types during GA iterations, we design the elitism as a matrix operation. Specifically, we consider the mutation matrix $M\_POP$ and the population matrix $POP$, selecting the dominant individuals according to the value calculated by the fitness function. The elitism operation can be expressed as follows:
\begin{equation}
    SI = \text{argsort}\left(\genfrac{[}{]}{0pt}{}{\text{F} \left(POP\right)}
    {\text{F} \left(M\_POP\right)}\right)
\end{equation}

\begin{equation}
\genfrac{[}{]}{0pt}{}{{N\_POP}_1}{{N\_POP}_2} = \text{sort}\left(\genfrac{[}{]}{0pt}{}{POP}{M\_POP}, \text{order}=SI\right)
\end{equation}
where $\text{argsort}\left(\cdot\right)$ returns the sorted index $SI$, arranging the fitness value matrix calculated by function $\text{F}(\cdot)$ in descending order. Using this index, $\genfrac{[}{]}{0pt}{}{POP}{M\_POP}\in \mathbb{R}^{2s \times 1}$ is sorted in row-major order, resulting in a combined sorted matrix. Then it is divided into two parts in row-major order: $N\_POP_1$ and $N\_POP_2$. Finally, the appropriate population sequence is chosen based on the evolutionary requirements of the algorithm. If the algorithm requires a population with high fitness, ${N\_POP}_2$ is retained; otherwise, ${N\_POP}_1$ is retained.

\subsection{Fitness Function Design}
\label{SixDST}

The fitness calculation operation of GA is a time-intensive operation, which significantly hampers the algorithm iteration efficiency. Although recent libraries like Evox and Evotorch offer general acceleration frameworks for evolutionary algorithms, they lack unified guidance on optimizing fitness functions. Specifically, we can decompose the GA-based PSSO problems fitness calculation into a perturbation update evaluation and a perturbation evaluation. Traditional GA-based PSSO methods typically update the perturbation using an index and evaluate it, which can be simplified as follows:

\begin{equation}
    \Delta f(\mathcal{G},p) = \mathcal{F}(\sum_{i=0}^{|p|
}\text{P}(\mathcal{G}, g_i))
\end{equation}
where $\text{P}(\cdot,\cdot)$ is the perturbation update function, the perturbation $p=[g_1, g_2, ..., g_k]$ consists of genes $g_i$ representing the $i$-th gene in the perturbation, $\mathcal{F}(\cdot)$ is the function that evaluates the impact of perturbation. However, these methods require numerous iterations, which leads to time-intensive calculations and inefficient use of available acceleration resources.

Based on the above analysis, we propose an optimization strategy that reconstructs the perturbation update and evaluation calculations, accelerating the calculation process and fully utilizing acceleration resources by converting the index-based iterative approach into matrix-based batch operations. The strategy can be summarized as follows:

\begin{equation}
    \Delta f(\mathcal{G},\mathit{p}) \approx \Delta \hat{f}(A,p) = \hat{\mathcal{F}}(\hat{\text{P}}(A, p))
\end{equation}
where $\hat{\mathcal{F}}(\cdot)$ is the matrix perturbation evaluation function used to assess disturbances, and $\hat{\text{P}}(\cdot,\cdot)$ is the perturbation update function designed to batch perturbations to $A$. It is worth noting that both node and edge perturbations are directly reflected in the adjacency matrix, with node deletion equivalent to removing all associated edges. Therefore, both types of perturbations can be treated as edge perturbations in adjacency matrix. Furthermore, we hope that the $\Delta \hat{f}(A,p)$ computed by $\hat{\mathcal{F}}(\cdot)$ is approximately equal to, or matches, the result of the traditional method, and the impact of this approximation on the perturbation evaluation is negligible. 



Specifically, to vividly and concretely describe our optimization strategy for fitness calculation, we use CND as an example and redesign the fitness function (SixDST) based on the Six Degrees of Separation Theorem \cite{travers1977experimental}. Firstly, convert the network into an adjacency matrix $A$, and apply the perturbation by $A=\hat{\text{P}}(A,p)$. $\hat{\text{P}}(\cdot,\cdot)$ can be easily implemented through batch calculations on matrices. $A$ contains node connection information, with higher-order connections obtained by multiplying itself. For example, elements in $A^2$ represent connections between nodes with a path length of 2. The accessibility matrix $M$ is computed by normalizing the cumulative matrix, obtained as the limit of the sum of $A_i$ as $N$ approaches infinity:

\begin{equation}
    \label{eq:normalizer}
    M=\text{Normalizer}\left(\lim_{N\rightarrow\infty}\sum_{i=1}^{N}A^i\right)
\end{equation}
The function $\text{Normalizer}(\cdot)$ sets each non-zero element in the cumulative matrix to 1, making $M$ indicate if a connection path of any length exists between nodes. However, this approach is computationally expensive.

To address this issue, we propose a simplification strategy based on the six degrees of separation theory \cite{travers1977experimental}, which states that social networks typically have a diameter of six or fewer. Thus, for networks with a diameter $\leq 6$, the parameter $N$ in \cref{eq:normalizer} is set to 6. Consequently, when the parameter $N$ satisfies $N\geq D(G)$, meaning that $N$ is greater than or equal to the target network diameter, \cref{eq:normalizer} simplifies to \cref{eq:normalizer2}:

\begin{equation}
    \label{eq:normalizer2}
    M=\text{Normalizer}\left(\sum_{i=1}^{N}A^i\right)
\end{equation}
Simplification reduces matrix power calculations, though the $N$-th power is still needed. To streamline further, the binomial theorem and matrix multiplication properties are used to derive a simplified expression:

\begin{equation}
    \left(A+I\right)^N=\sum_{r=0}^{N}C_N^rA^{N-r}I^r=\sum_{r=0}^{N}k_rA^{N-r}
\end{equation}
\begin{equation}
    \label{eq:normalizer3}
    M=\resizebox{0.78\hsize}{!}{$\text{Normalizer}\left(\sum_{i=1}^{N}A^i\right)=\text{Normalizer}\left(\left(A+I\right)^N\right)$}
\end{equation}
where $I$ is the identity matrix, $k_r = C_N^r$ is a non-zero parameter, and its normalization can be ignored. Therefore, the accessibility matrix $M$ can be calculated as shown in \cref{eq:normalizer3}. Additionally, by leveraging matrix self-multiplication, the number of matrix multiplications required for $(A+I)^N$ can be reduced. Specifically, $(A+I)(A+I)^2(A+I)^4\cdots$ are computed through matrix multiplications, until the parameter $N$ satisfies $N\geq D(G)$. After that, the time complexity for the computation of the accessibility matrix is reduced from $\mathcal{O}(N\cdot n^3)$ to $\mathcal{O}(\log_2{N}\cdot n^3)$. Each row of the accessibility matrix $M$ shows a node's connections, with non-zero elements indicating its connected component size. Summing these values gives each node's connectivity, with the largest representing maximum connectivity number(MCN), making this operation suitable as a fitness function for the CND task.

\subsection{Acceleration Mode Design}

GA performs independent mutations and fitness calculations for each individual in the population, and these operations can be decomposed into many small tasks, making them highly suitable for parallel computing. Additionally, with the continuous growth of hardware resources, distributed acceleration has become a practical approach to enhance computational efficiency. To address this, we divide the calculation process of GA into global and fitness calculations and propose four acceleration modes to cover these calculation processes fully. The baseline mode is the original algorithm running in a CPU environment, labeled as \textbf{CPU} mode. The other four acceleration modes are as follows:





\textbf{Single process in genetic operation}(S):
Considering the limited acceleration resources, we propose the S mode. In this mode, the fitness calculation and other genetic operators are executed on one GPU, effectively demonstrating the parallel acceleration advantage that GAPA can provide on limited acceleration resources. The workflow is illustrated in \cref{fig:Main Architecture}.


\begin{figure}[!h]
\centering
\setlength{\abovecaptionskip}{-0.2cm}
\includegraphics[width=0.75\linewidth]{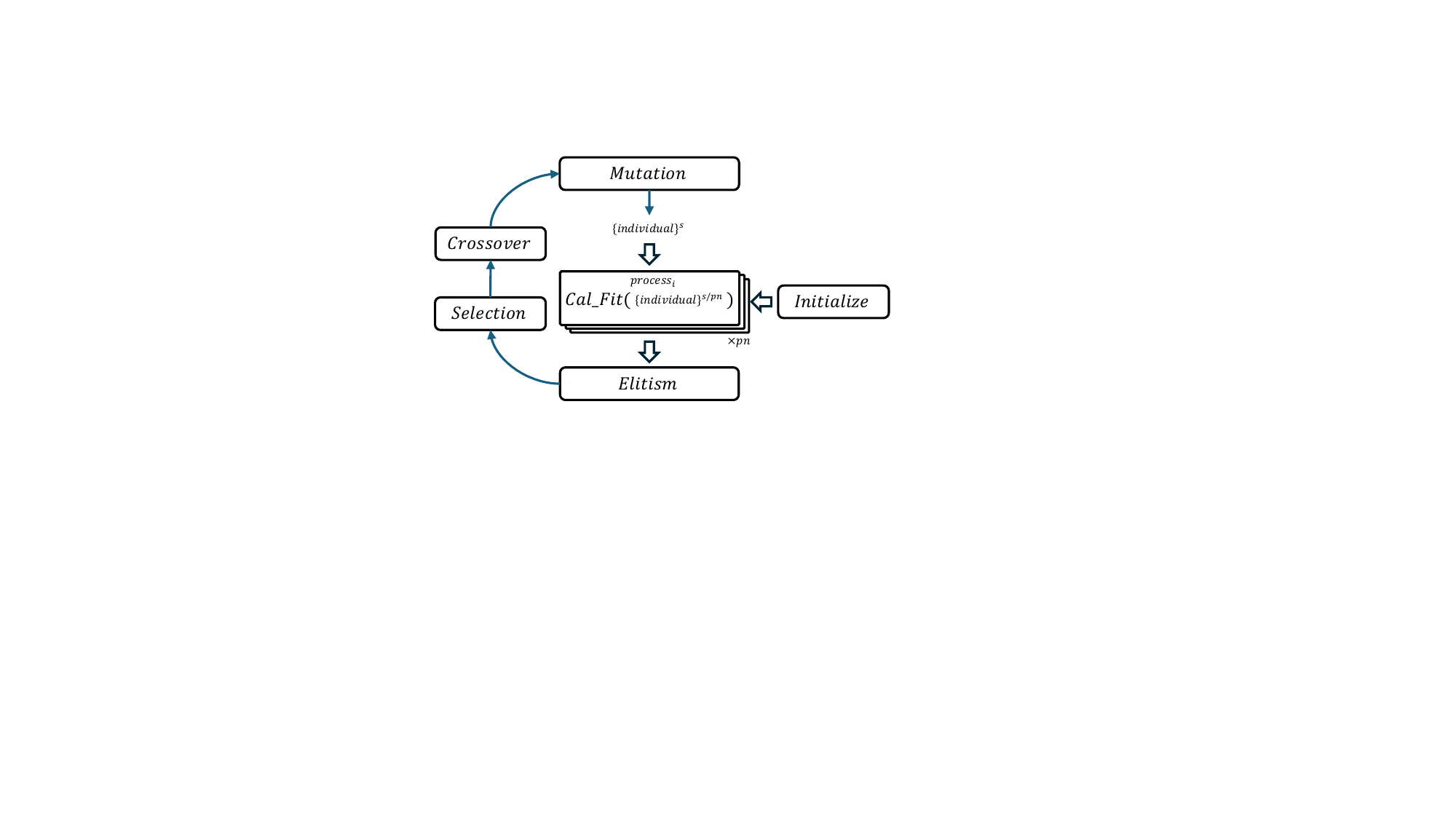}\\\
\caption{Workflow of SM mode acceleration. Specifically, in this mode, ${individual}^{s}$ is evenly distributed and assigned to $pn$ processes to calculate fitness.}


\label{fig:SM}
\end{figure}

\textbf{Single process in genetic operation \& Multi-process in fitness calculation}(SM): 
Fitness calculation is the most time-intensive part of the GA. To accelerate this operation, we proposed the SM mode, an extension of the S mode. This mode extends the fitness calculation to a distributed acceleration environment, further improving the iteration efficiency of the GA. As shown in \cref{fig:SM}, the fitness calculation operation is evenly distributed across each GPU, fully leveraging the advantages of distributed computing. Specifically, \textit{Initialize} generates the initial population, completing the \textit{Cal\_Fit} and \textit{Elitism} operations. The elite individuals are then sent to \textit{Selection} to begin the next iteration. After that, The population undergoes \textit{Crossover} and \textit{Mutation}, resulting in the mutation population ${individual}^{s}$. This population is divided into ${individual}^{s/pn}$ based on the number of processes $pn$, with \textit{Cal\_Fit} applied to each subset. Finally, the fitness values and populations from each process are combined in sequence, which is sent to \textit{Elitism} to complete the iteration.

\begin{figure}[!t]
\centering
\setlength{\abovecaptionskip}{-0.2cm}
\includegraphics[width=0.80\linewidth]{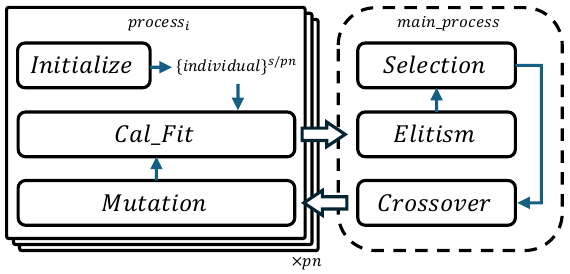}\\\
\caption{Workflow of M mode acceleration. Specifically, in this mode, the population undergoes the operators of \textit{Initialize}, \textit{Mutation}, and \textit{Cal\_Fit} through separate processes. $main\_process$ collects data from the $pn$ processes, and redistributes it back after the operators of \textit{Elitism}, \textit{Selection}, and \textit{Crossover}.}


\label{fig:MS}
\end{figure}

\textbf{Multi-process in genetic operation}(M): 
In the SM mode, the distributed acceleration of fitness calculation involves creating and destroying numerous processes, which can significantly increase the iteration cost of the GA. Additionally, initialization, mutation, and fitness calculation operations in genetic algorithms, depending on a single individual, are suitable for distributed acceleration. However, selection, crossover, and elitism operations, depending on the entire population, are not ideal for such distributed strategies. To address these issues, we propose the M mode, which designs the entire GA as a multi-process structure. The accelerated workflow is shown in \cref{fig:MS}, where each process only exchanges data when necessary. Compared to the SM mode, the M mode minimizes the overhead of process creation and destruction by limiting it to the number of GPUs used, thereby reducing the iteration cost of the GA. Specifically, \textit{Initialize} generates the initial population ${individual}^{s/pn}$ in each process, completing the \textit{Cal\_Fit} and \textit{Elitism} operations. The elite individuals are then sent to \textit{Selection} to begin the next iteration. Secondly, the $main\_process$ collects data from the $pn$ processes, redistributes it via \textit{Elitism}, \textit{Selection}, and \textit{Crossover}, and sends the results back. After that, each process generates the corresponding population through \textit{Mutation} and \textit{Cal\_Fit}. Finally, the fitness values and populations from each process are combined in sequence, which is sent to \textit{Elitism} to complete the iteration.



\begin{figure}[!h]
\centering
\setlength{\abovecaptionskip}{-0.2cm}
\includegraphics[width=0.80\linewidth]{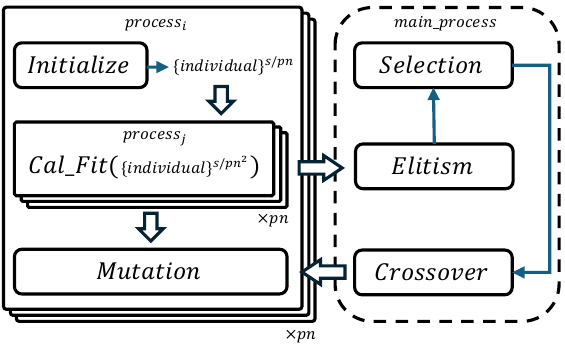}\\\
\caption{Workflow of MNM mode acceleration. Specifically, this mode extends the M mode by adding distributed acceleration to the fitness calculation, which follows the same distributed acceleration approach as the SM mode.}
\label{fig:MM}
\end{figure}

\textbf{Multi-process in genetic operation \& Nested Multi-process fitness calculation}(MNM): 
Building upon the M mode, we further explore the distributed acceleration of the fitness calculation by leveraging the global multi-process structure to maximize the use of acceleration resources, leading to the proposal of the MNM mode. The accelerated workflow is shown in \cref{fig:MM}. In this mode, in addition to the multi-processing of the genetic operation, the fitness calculation is also distributed and accelerated within each process. Specifically, the MNM mode extends the M mode by adding distributed acceleration to the fitness calculation, following the same approach as the SM mode.

\begin{figure}[!h]
\centering
\setlength{\abovecaptionskip}{-0.2cm}
\includegraphics[width=0.80\linewidth]{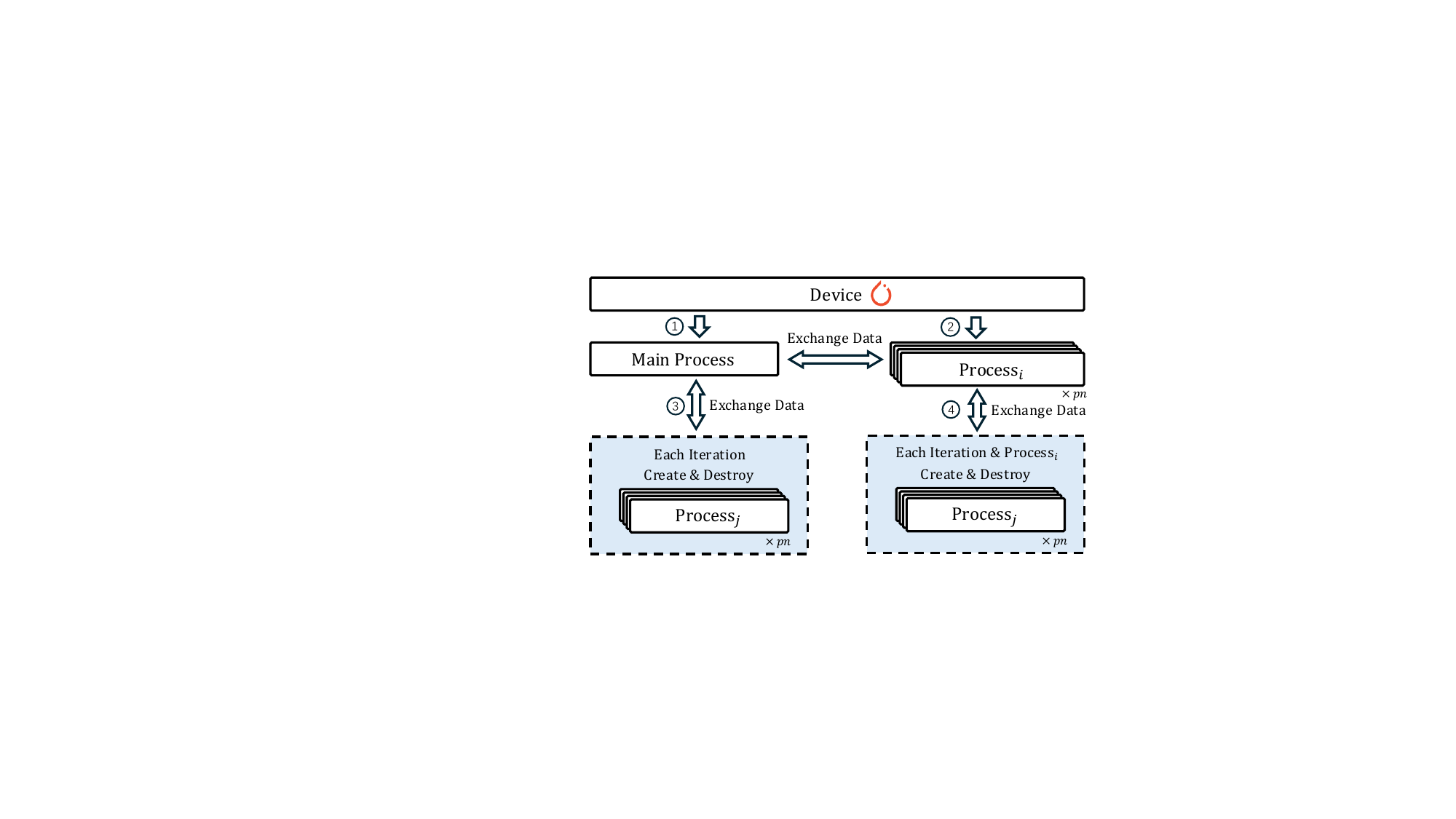}\\\
\caption{Illustration of the additional computation time introduced by multiple processes. This extra time includes the data exchange between processes and the creation and destruction of processes during each iteration. Specifically, \raisebox{0.5pt}{\textcircled{\raisebox{-0.5pt} {1}}} represents the S mode, \raisebox{0.5pt}{\textcircled{\raisebox{-0.5pt} {1}}}\raisebox{0.5pt}{\textcircled{\raisebox{-0.5pt} {3}}} represents the SM mode, \raisebox{0.5pt}{\textcircled{\raisebox{-0.5pt} {1}}}\raisebox{0.5pt}{\textcircled{\raisebox{-0.5pt} {2}}} represents the M mode, and \raisebox{0.5pt}{\textcircled{\raisebox{-0.5pt} {1}}}\raisebox{0.5pt}{\textcircled{\raisebox{-0.5pt} {2}}}\raisebox{0.5pt}{\textcircled{\raisebox{-0.5pt} {3}}}\raisebox{0.5pt}{\textcircled{\raisebox{-0.5pt} {4}}} represents the MNM mode.}
\label{fig:Process}
\end{figure}

Finally, \cref{fig:Process} illustrates the additional computational time introduced by executing multiple processes across different acceleration modes. This additional computation time primarily arises from data exchange between processes and the creation and destruction of processes during each iteration. Specifically, the SM mode incurs significant additional computation time due to frequent process creation and destruction; the M mode is mainly affected by data exchange between processes, leading to increased additional computation time; while the MNM mode experiences all additional time.


\begin{table}[!t]
    \renewcommand{\arraystretch}{1.2}
    \centering
    \caption{Task Categories along with Their Corresponding Algorithms and Datasets. The number of network nodes and edges is represented as Node and Edge. Avg.Degree represents the average degree of the network.}
    
    \label{table:DA}
    \resizebox{1\linewidth}{!}{
    \begin{tabular}{c|c|c|ccc}
    \hline\hline
    \textbf{Task} & \textbf{Algorithm} & \textbf{Dataset} & \textbf{Node} & \textbf{Edge} & \textbf{Avg.Degree} 
    \\ 
    \hline
    
    \multirow{4}{*}{CDA}  
    & \multirow{4}{*}{
    \renewcommand{\arraystretch}{1.5}
    \begin{tabular}[c]{@{}c@{}}
        QAttack \\ CDA-EDA \\ CGN
    \end{tabular}
    } 
    & Karate & 34 & 78 & 4.58 \\
    & & Dolphins & 62 & 159 & 5.12 \\
    & & Football & 115 & 613 & 10.66 \\
    & & Email-EU-Core(EEC1000) & 1005 & 16706 & 33.25 \\
    \hline
    \hline
    
    \multirow{4}{*}{CND} 
    & \multirow{4}{*}{
    \renewcommand{\arraystretch}{1.5}
    \begin{tabular}[c]{@{}c@{}}
        CutOff \\ SixDST \\ TDE
    \end{tabular}
    } 
    & ForestFire(FF500) & 500 & 828 & 3.31 \\
    & & Erdos-Renyi(ER500) & 466 & 700 & 3.00 \\
    & & Barabasi-Albert(BA500) & 500 & 499 & 2.00 \\
    & & Yeast1 & 2018 & 2705  & 2.60 \\ 
    \hline
    \hline

    \multirow{4}{*}{NCA} 
    & \multirow{4}{*}{
    \renewcommand{\arraystretch}{1.5}
    \begin{tabular}[c]{@{}c@{}}
        NCA-GA \\ GANI 
    \end{tabular}
    } 
    & Chameleon\_Filtered(CF900) & 890 & 8854 & 19.89 \\
    & & Squirrel\_Filtered(SF2000) & 2223 & 46998 & 42.28 \\
    & & Cora & 2485 & 7554  & 6.08 \\
    & & Citeseer & 3327 & 4552  & 2.74 \\ 
    \hline
    \hline
    
    \multirow{4}{*}{LPA} 
    & \multirow{4}{*}{
    \renewcommand{\arraystretch}{1.5}
    \begin{tabular}[c]{@{}c@{}}
        LPA-GA \\ LPA-EDA 
    \end{tabular}
    }
    & Dolphins & 62 & 159 & 5.12 \\
    & & Jazz & 255 & 2742 & 27.69 \\
    & & USAir97 & 332 & 2126 & 12.81 \\
    & & HumanDiseasome(HD500) & 516 & 1188 & 4.60 \\ 
    \hline
    \hline
    
    
    \end{tabular}
    }
\end{table}

\begin{table*}[!t]
\centering
\renewcommand{\arraystretch}{1.2}
\vspace{-0.3cm}
\caption{Experimental parameter configuration of different algorithms}
\label{tab: Params}
\begin{tabular}{c|cccccccccc}
    \hline\hline
    \textbf{Parameter} & QAttack & CDA-EDA & CGN  & CutOff & SixDST & TDE  & NCA-GA & GANI & LPA-GA & LPA-EDA \\
    \hline
    $pc$ & 0.8 & 0.6 & 0.7 & 0.6 & 0.5 & 0.5 & 0.7 & 0.7 & 0.7 & - \\
    
    $pm$ & 0.1 & 0.2 & 0.01 & 0.2 & 0.3 & 0.3 & 0.3 & 0.3 & 0.1 & 0.1 \\

    $eda$ & - & - & - & - & - & - & - & - & - & 50 \\

    iteration & 1500 & 1500 & 1500 & 5000 & 5000 & 5000 & 1500 & 1500 & 500 & 500 \\

    $pop\_size$ & 100 & 100 & 100 & 80 & 80 & 10 & 80 & 80 & 50 & 50 \\
    
    perturbation rate & 0.1 & 0.1 & 0.1 & 0.1 & 0.1 & 0.1 & 0.025 & 0.05 & 0.1 & 0.1 \\
    \hline\hline
\end{tabular}
\vspace{-0.3cm}
\end{table*}

\section{Experiment}
\label{Experiment}

This section conducts extensive experiments from the perspectives of acceleration mode, population size, distributed acceleration, and comparative analysis, which comprehensively evaluate GAPA and validate its effectiveness.


\subsection{Datasets and Algorithms}
\label{DA}

To verify the performance of GAPA, 12 widely used dataset are utilized in the experiment with four tasks, which include \textbf{CDA}(Karate\cite{zachary1977information}, Dolphins\cite{lusseau2003bottlenose}, Football\cite{girvan2002community}, EEC1000\cite{leskovec2007graph}), \textbf{CND}(FF500\cite{ff}, ER500\cite{er}, BA500\cite{ba}, Yeast1\cite{zhou2020variable}), \textbf{NCA}(CF900\cite{chameleon4squirrel}, SF2000\cite{chameleon4squirrel}, Cora\cite{cora}, Citeseer\cite{citeseer}) and \textbf{LPA}(Dolphins, Jazz\cite{gleiser2003community}, USAir97\cite{zhou2020variable}, HD500\cite{zhou2020variable}). We also implement our framework with 10 algoriths including QAttack\cite{8714065}, CDA-EDA\cite{9172881}, CGN\cite{cgn}, CutOff\cite{cutoff}, SixDST, TDE\cite{tde}, NCA-GA\cite{dai2018adversarial} GANI\cite{gani}, LPA-GA\cite{lpa-eda}, and LPA-EDA\cite{lpa-eda}. The statistics of the datasets and algorithms are summarized in~\cref{table:DA}.

\begin{table*}[!t]
    \renewcommand{\arraystretch}{1.2}
    \centering
    \vspace{-0.12cm}
    \caption{Performance Comparison of 4 Tasks Across Various Datasets and Acceleration Modes. The best and sub-optimal are highlighted in bold and underlined, respectively. Times that take more than twelve hours are marked using $'/'$ and total computation times (Time) in seconds.}
\label{table:Whole Table}
    \renewcommand\arraystretch{1.5}
    \tabcolsep=0.6mm
    
    \resizebox{1\linewidth}{!}{
    \begin{tabular}{c|c ccccc|ccccc|ccccc|ccccc}
    
        \hline\hline
    %
    %
        \multicolumn{22}{c}{\textbf{CDA}}
        \\
        \hline
        
        \multirow{2}{*}{\textbf{Method}} 
        & \multirow{2}{*}{\textbf{Metric}} 
        & \multicolumn{5}{c|}{Karate} & \multicolumn{5}{c|}{Dolphins} 
        & \multicolumn{5}{c|}{Football} & \multicolumn{5}{c}{EEC1000} 
        \\ 
        \cline{3-22} 
        
        & & CPU & S & SM & M & MNM 
        & CPU & S & SM & M & MNM 
        & CPU & S & SM & M & MNM 
        & CPU & S & SM & M & MNM 
        \\ 
        \hline

        \rowcolor{gray!20}
        \multirow{3}{*}{CDA-EDA} 
        \cellcolor{white} & \multicolumn{1}{|c}{Time} & \underline{295.64} & 386.79 & 339.96 & \textbf{230.86} & 417.33
        & 947.72 & 397.48 & \underline{371.88} & \textbf{234.73} & 430.55
        & 2799.76 & 511.89 & 479.95 & \textbf{293.20} & \underline{453.70}
        & / & 7554.89 & 9523.55 & \textbf{3622.62} & \underline{4190.80}
        \\ 
        
        & Q & 0.39 & 0.38 & 0.40 & 0.40 & 0.39
        & 0.50 & 0.51 & 0.51 & 0.51 & 0.50
        & 0.61 & 0.59 & 0.59 & 0.60 & 0.60
        & / & 0.42 & 0.43 & 0.42 & 0.40
        \\ 
        
        & NMI & 0.75 & 0.60 & 0.89 & 0.81 & 0.89
        & 0.77 & 0.96 & 0.74 & 0.92 & 0.88
        & 0.94 & 0.94 & 0.89 & 1.00 & 0.95
        & / & 0.74 & 0.84 & 0.80 & 0.89
        \\ 
        \hline

        \rowcolor{gray!20}
        \multirow{3}{*}{CGN} 
        \cellcolor{white} & \multicolumn{1}{|c}{Time} & 242.15 & \underline{141.06} & 224.26 & \textbf{101.99} & 219.53
        & 282.83 & \underline{183.20} & 281.88 & \textbf{122.08} & 238.74
        & 404.52 & 350.64 & 467.23 & \textbf{205.15} & \underline{335.84}
        & 13920.01 & 6574.48 & 12217.13 & \textbf{3784.77} & \underline{6293.86}
        \\ 
        
        & Q & 0.39 & 0.39 & 0.36 & 0.39 & 0.38
        & 0.50 & 0.49 & 0.50 & 0.50 & 0.50 
        & 0.60 & 0.59 & 0.59 & 0.59 & 0.59
        & 0.43 & 0.43 & 0.43 & 0.43 & 0.43    
        \\ 
        
        & NMI & 0.40 & 0.37 & 0.34 & 0.34 & 0.34
        & 0.42 & 0.41 & 0.49 & 0.40 & 0.48
        & 0.83 & 0.79 & 0.79 & 0.80 & 0.77
        & 0.62 & 0.62 & 0.64 & 0.65 & 0.59    
        \\ 
        
        \hline
        \rowcolor{gray!20}
        \multirow{3}{*}{QAttack}
        \cellcolor{white} & \multicolumn{1}{|c}{Time} & \underline{136.07} & 153.26 & 275.58 & \textbf{107.58} & 200.08 
        & 231.67 & \underline{227.92} & 371.73 & \textbf{144.25} & 236.45 
        & 538.37 & 448.62 & 644.43 & \textbf{255.15} & \underline{400.11} 
        & 14448.30 & 6997.85 & 12246.61 & \textbf{3560.23} & \underline{5703.01}
        \\

        & Q & 0.30 & 0.29 & 0.29 & 0.31 & 0.29
        & 0.41 & 0.39 & 0.41 & 0.41 & 0.41
        & 0.56 & 0.55 & 0.55  & 0.56 & 0.55
        & 0.39 & 0.40 & 0.40 & 0.39 & 0.40
        \\ 

        & NMI & 0.92 & 0.87 & 0.58 & 0.40 & 0.38
        & 0.64 & 0.72 & 0.89 & 0.56 & 0.55 
        & 1.00 & 0.92 & 0.95 & 0.86 & 0.88 
        & 0.81 & 0.78 & 0.79 & 0.68 & 0.69
        \\ 
        
        \hline\hline

    %
    %
        \multicolumn{22}{c}{\textbf{CND}}
        \\
        \hline
        
        \multirow{2}{*}{\textbf{Method}}
        & \multirow{2}{*}{\textbf{Metric}} 
        & \multicolumn{5}{c|}{FF500} & \multicolumn{5}{c|}{ER500} 
        & \multicolumn{5}{c|}{BA500} & \multicolumn{5}{c}{Yeast1}      
        \\ 
        \cline{3-22} 
        
        & & CPU & S & SM & M & MNM 
        & CPU & S & SM & M & MNM 
        & CPU & S & SM & M & MNM 
        & CPU & S & SM & M & MNM      
        \\ 
        \hline

        \rowcolor{gray!20}
        \multirow{3}{*}{TDE}
        \cellcolor{white} & \multicolumn{1}{|c}{Time} & 9176.49 & 2015.60 & 2987.54 & \textbf{1046.37} & \underline{1496.07}
        & 8033.15 & 1830.29 & 2225.43 & \textbf{959.34} & \underline{1287.88}
        & 8896.82 & 1902.27 & 2269.46 & \textbf{989.63}  & \underline{1310.84}
        & / & 26014.88 & 25807.94 & \underline{13317.74} & \textbf{13101.58}
        \\ 
        
        & PC(G) & 8461 & 4315 & 5213 & 3657 & 3643
        & 69044 & 69792 & 69048 & 70151 & 65807
        & 739 & 804 & 823 & 741 & 954
        & / & 570407 & 636283 & 584342 & 583208       
        \\ 
        
        & MCN & 66 & 44 & 63 & 42 & 40
        & 372 & 374 & 372 & 375 & 363 
        & 15 & 13 & 16 & 15 & 15
        & / & 1068 & 1128 & 1081 & 1080

        \\ 
        \hline

        \rowcolor{gray!20}
        \multirow{3}{*}{CutOff}  
        \cellcolor{white} & \multicolumn{1}{|c}{Time} & 2264.56 & 992.36 & 1347.50 & \textbf{814.20} & \underline{984.94}
        & 1994.55 & \underline{932.07} & 1224.60 & \textbf{782.32} & 942.27
        & 1789.62 & 1025.36 & 1323.35 & \textbf{804.28} & \underline{977.18}
        & 6809.12 & 2119.65 & 2697.05 & \textbf{1391.76} & \underline{1745.70}
        \\ 
        
        & PC(G) & 2357 & 2129 & 2355 & 2188 & 2391
        & 55793 & 54441 & 51901 & 55109 & 55704
        & 439 & 431 & 457 & 396 & 429 
        & 258955 & 279325 & 236480 & 298771 &  275299
        \\ 
        
        & MCN & 27 & 28 & 31 & 24 & 27
        & 334 & 330 & 322 & 332 & 333
        & 9 & 8 & 9 & 7 & 9
        & 718 & 745 & 668 & 772 & 741     
        \\ 
        \hline

        \rowcolor{gray!20}
        \multirow{3}{*}{SixDST} 
        \cellcolor{white} & \multicolumn{1}{|c}{Time} & 2906.85 & \textbf{147.71} & 188.00 & \underline{183.23} & 806.04
        & 2532.48 & \textbf{148.08} & 197.32 & \underline{182.75} & 800.62
        & 2951.43 & \textbf{149.62} & 203.01 & \underline{181.10} & 805.78
        & / & 1454.63 & \underline{834.40} & \textbf{824.97} & 1550.85 
        \\ 
        
        & PC(G) & 1883 & 2002 & 1899 & 2041 & 2049 
        & 72106 & 48132 & 50290 & 50877 & 47755
        & 416 & 483 & 430 & 516 & 443
        & / & 348693 & 335321 & 150339 & 263626
        \\ 
        
        & MCN & 18 & 19 & 17 & 18 & 18
        & 156 & 310 & 317 & 319 & 309
        & 7 & 7 & 7 & 8  & 7
        & / & 229 & 174 & 177 & 170

        \\ 
        \hline\hline

    %
    %
        \multicolumn{22}{c}{\textbf{NCA}}
        \\
        \hline
        
        \multirow{2}{*}{\textbf{Method}}  
        & \multirow{2}{*}{\textbf{Metric}}
        & \multicolumn{5}{c|}{CF900} & \multicolumn{5}{c|}{SF2000}  
        & \multicolumn{5}{c|}{Cora} & \multicolumn{5}{c}{Citeseer}               
        \\ 
        \cline{3-22} 
        
        & & CPU & S & SM & M &  MNM 
        & CPU & S & SM & M &  MNM 
        & CPU & S & SM & M &  MNM 
        & CPU & S & SM & M &  MNM 
        \\ 
        \hline

        \rowcolor{gray!20}
        \multirow{3}{*}{GANI} 
        \cellcolor{white} & \multicolumn{1}{|c}{Time} & 5735.29 & 1502.38 & \underline{1432.17} & \textbf{1267.73} & 2745.32
        & 29678.14 & 2289.63 & \underline{1801.83} & \textbf{1801.49} & 3637.66
        & 21957.48 & 5488.61 & \underline{4978.91} & \textbf{4559.15} & 9116.75
        & 13468.55 & 5218.17 & \underline{5123.09} & \textbf{4693.43} & 8463.04
        \\ 
        
        & Accuracy & 0.36 & 0.36 & 0.34 & 0.40 & 0.38
        & 0.37 & 0.37 & 0.37 & 0.36 & 0.37 
        & 0.74 & 0.68 & 0.61 & 0.65 & 0.63 
        & 0.60 & 0.59 & 0.60 & 0.59 & 0.59
        \\ 
        
        & ASR & 14.43 & 27.83 & 24.74 & 17.52 & 19.58 
        & 16.82 & 15.04 & 16.59 & 15.92 & 23.45 
        & 9.15 & 23.28 & 30.72 & 25.44 & 28.46
        & 14.32 & 25.91 & 23.65 & 26.92 & 23.80
        \\ 
        \hline

        \rowcolor{gray!20}
        \multirow{3}{*}{NCA-GA} 
        \cellcolor{white} & \multicolumn{1}{|c}{Time} & 685.96 & \underline{284.11} & 349.79 & \textbf{178.24} & 504.56
        & 9541.54 & 1333.31 & 1406.58 & \textbf{716.19} & \underline{1106.64 }
        & 665.57 & \underline{268.23} & 362.18 & \textbf{149.93} & 487.12
        & 835.51 & \underline{205.01} & 338.08 & \textbf{156.97} & 528.42  
        \\ 
        
        & Accuracy & 0.41 & 0.41 & 0.41 & 0.41 & 0.42
        & 0.37 & 0.37 & 0.37 & 0.37 & 0.36
        & 0.81 & 0.81 & 0.81 & 0.81 & 0.81
        & 0.72 & 0.72 & 0.71 & 0.71 & 0.72
        \\ 
        
        & ASR & 3.09 & 4.64 & 3.61 & 4.12 & 1.54
        & 1.54 & 3.31 & 1.76 & 2.88 & 2.43
        & 1.05 & 1.05 & 1.16 & 1.05 & 1.26
        & 1.70 & 1.10 & 1.00 & 1.49 & 0.80
        \\ 
        
        \hline\hline

    %
    %
        \multicolumn{22}{c}{\textbf{LPA}}
        \\
        \hline
        
        \multirow{2}{*}{\textbf{Method}}  
        & \multirow{2}{*}{\textbf{Metric}}
        & \multicolumn{5}{c|}{Dolphins} & \multicolumn{5}{c|}{HD500} & \multicolumn{5}{c|}{Jazz}  
        & \multicolumn{5}{c}{USAir97}               
        \\ 
        \cline{3-22} 
        
        & & CPU & S & SM & M &  MNM 
        & CPU & S & SM & M &  MNM 
        & CPU & S & SM & M &  MNM 
        & CPU & S & SM & M &  MNM 
        \\ 
        \hline

        \rowcolor{gray!20}
        \multirow{3}{*}{LPA-EDA} 
        \cellcolor{white} & \multicolumn{1}{|c}{Time} & \underline{105.17} & 134.54 & 187.40 & \textbf{79.11} & 118.90
        & 1290.77 & 2759.25 & 3508.31 & \textbf{1056.93} & \underline{1180.07}
        & 4949.55 & 3416.70 & 3723.08 & \underline{1824.89} & \textbf{1804.14}
        & 8243.65 & 3204.75 & 3648.44 & \textbf{1478.93} & \underline{1733.58}
        \\ 
        
        & Precision & 0.00 & 0.00 & 0.00 & 0.00 & 0.00
        & 0.26 & 0.27 & 0.23 & 0.22
        & 0.13 & 0.12 & 0.12 & 0.11 & 0.12 
        & 0.31 & 0.31 & 0.33 & 0.30 & 0.32 & 0.36
        \\ 
        
        & AUC & 0.60 & 0.51 & 0.50 & 0.56 & 0.57 
        & 0.85 & 0.78 & 0.81 & 0.797 & 0.80
        & 0.76 & 0.67 & 0.67 & 0.69 & 0.66 
        & 0.92 & 0.91 & 0.91 & 0.90 & 0.91
        \\ 
        \hline

        \rowcolor{gray!20}
        \multirow{3}{*}{LPA-GA} 
        \cellcolor{white} & \multicolumn{1}{|c}{Time} & \textbf{47.88} & 76.58 & 141.18 & \underline{55.87} & 83.60
        & \textbf{539.81} & 1182.81 & 1527.95 & \underline{654.66} & 762.73
        & 1728.45 & 1120.56 & 1289.60 & \textbf{573.51} & \underline{678.08} 
        & 2506.87 & 1818.61 & 1898.27 & \textbf{944.04} & \underline{1012.57}
        \\ 
        
        & Precision & 0.00 & 0.03 & 0.00 & 0.00 & 0.00
        & 0.39 & 0.29 & 0.26 & 0.27 & 0.26
        & 0.16 & 0.15 & 0.15 & 0.13 & 0.10
        & 0.38 & 0.32 & 0.34 & 0.31 & 0.30
        \\ 
        
        & AUC & 0.63 & 0.55 & 0.52 & 0.56 & 0.63
        & 0.89 & 0.83 & 0.79 & 0.82 & 0.82
        & 0.80 & 0.70 & 0.70 & 0.72 & 0.69
        & 0.93 & 0.90 & 0.89 & 0.93 & 0.90
        
        \\ 

        \hline\hline
        
    \end{tabular}
    }

\end{table*}

\subsection{Experimental Setups}
\label{ESP}

To ensure a fair comparison, the main experiments in this paper are conducted on Intel(R) Xeon(R) Gold 5218R CPU @ 2.10GHz with two NVIDIA A100 SXM4 40GB GPUs. Considering the need for distributed acceleration, the distributed acceleration and acceleration framework comparison experiment is conducted on Intel(R) Xeon(R) Gold 5218R CPU @ 2.10GHz with six Tesla V100 SXM2 16GB GPUs. Moreover, for different algorithms, we follow the parameter settings of the original paper, where the specific settings are shown in~\cref{tab: Params}.

\begin{table*}[]
\renewcommand{\arraystretch}{1.2}
\centering
\caption{Comparison of computation time of 4 tasks at different population sizes across different datasets and acceleration modes. The best and sub-optimal are highlighted in bold and underlined, respectively. Times that take more than twelve hours are marked using $'/'$ and total computation times (Time) in seconds. In CND task, TDE uses the population size specified in $'()'$.}
\label{table:Scale}
\renewcommand\arraystretch{1.5}
\tabcolsep=0.6mm

\resizebox{1\linewidth}{!}{
\begin{tabular}{c|ccccccc|cccccc|cccccc|cccccc}
\hline\hline

    \multicolumn{26}{c}{\textbf{CDA}} \\
    \hline
    
    \multirow{2}{*}{\textbf{Method}}  & \multirow{2}{*}{\textbf{Mode}} & \multicolumn{6}{c|}{Karate} & \multicolumn{6}{c|}{Dolphins} & \multicolumn{6}{c|}{Football} & \multicolumn{6}{c}{EEC1000} \\
    \cline{3-26}
    
    & & 20 & 40 & 60 & 80 & 100 & 120 & 20 & 40 & 60 & 80 & 100 & 120 & 20 & 40 & 60 & 80 & 100 & 120 & 20 & 40 & 60 & 80 & 100 & 120 \\
    \hline

    \multirow{3}{*}{CDA-EDA} & CPU & \textbf{67.5} & \underline{147.4} & \underline{188.5} & \underline{248.7} & \underline{295.6} & \underline{349.6} & 207.5 & 422.5 & 618.7 & 815.7 & 947.7 & 1240.4 & 549.1 & 1108.4 & 1706.4 & 2159.3 & 2799.8 & 3404.8 & / & / & / & / & / & / \\
    
    & S & \underline{82.0} & 152.6 & 232.5 & 307.5 & 386.8 & 472.1 & \underline{86.2}    & \underline{172.3}    & \underline{247.4}    & \underline{341.1}    & \underline{397.5}    & \underline{505.7}    & \underline{106.1}   & \underline{211.7}    & \underline{307.3}    & \underline{409.5}    & \underline{511.9}    & \underline{625.0}    & \underline{1468.6}   & \underline{2870.4}    & \underline{4284.9}    & \underline{5719.4}    & \underline{7554.9}    & \underline{8601.6}    \\

    & M & 82.7 & \textbf{116.3} & \textbf{172.5} & \textbf{200.6} & \textbf{230.9} & \textbf{274.2} & \textbf{78.5} & \textbf{127.9} & \textbf{172.4} & \textbf{204.8} & \textbf{234.7} & \textbf{290.5} & \textbf{92.9} & \textbf{146.9} & \textbf{188.4} & \textbf{242.4} & \textbf{293.2} & \textbf{356.8} & \textbf{876.9} & \textbf{1614.9} & \textbf{2397.3} & \textbf{3134.4} & \textbf{3622.6} & \textbf{4707.5} \\

    \hline

    \multirow{3}{*}{CGN} & CPU & \underline{46.9}    & 92.7          & 137.3         & 189.6         & 242.1          & 278.9          & 55.1          & 109.0         & 163.3         & 218.8          & 282.8          & 325.8          & 77.4          & 154.1          & 232.0          & 309.0          & 404.5          & 461.4          & 2721.8         & 5525.6          & 8038.8          & 10895.7         & 13920.0         & 16347.9         \\
    
    & S & \textbf{31.1} & \textbf{60.6} & \underline{87.1}    & \underline{127.0}   & \underline{141.1}    & \underline{175.2}    & \textbf{39.8} & \underline{77.1}    & \underline{119.8}   & \underline{149.5}    & \underline{183.2}    & \underline{217.6}    & \textbf{72.5} & \underline{140.7}    & \underline{212.9}    & \underline{284.1}    & \underline{350.6}    & \underline{411.7}    & \underline{1440.0}   & \underline{2785.4}    & \underline{4157.0}    & \underline{5406.2}    & \underline{6574.5}    & \underline{8123.4}    \\

    & M & 50.1          & \underline{63.2}    & \textbf{73.0} & \textbf{91.4} & \textbf{102.0} & \textbf{113.3} & \underline{52.8}    & \textbf{71.6} & \textbf{85.9} & \textbf{103.1} & \textbf{122.1} & \textbf{140.2} & \underline{74.0}    & \textbf{111.9} & \textbf{138.2} & \textbf{172.6} & \textbf{205.2} & \textbf{236.2} & \textbf{932.1} & \textbf{1696.3} & \textbf{2454.2} & \textbf{3215.0} & \textbf{3784.8} & \textbf{4701.3} \\

    \hline   

    \multirow{3}{*}{QAttack} & CPU & \textbf{26.2} & \textbf{51.2} & \textbf{77.0} & \underline{102.4}   & \underline{136.1}    & \underline{153.2}    & \textbf{43.8} & \underline{85.7}    & \underline{129.6}    & \underline{173.2}    & \underline{231.7}    & \underline{259.0}    & 102.9         & 205.2          & 307.3          & 411.6          & 538.4          & 615.5          & 2813.9         & 5627.7          & 8378.2          & 11190.0         & 14448.3         & 16904.4         \\
    
    & S & \underline{37.0}    & \underline{66.3}    & 99.5          & 130.6         & 153.3          & 184.4          & \underline{51.8}    & 96.3          & 154.7          & 197.0          & 227.9          & 270.9          & \underline{98.8}    & \underline{195.7}    & \underline{273.4}    & \underline{373.1}    & \underline{448.6}    & \underline{582.3}    & \underline{1447.0}   & \underline{2800.3}    & \underline{4224.1}    & \underline{5610.5}    & \underline{6997.8}    & \underline{8268.2}    \\

    & M & 51.5          & 66.5          & \underline{82.9}    & \textbf{98.2} & \textbf{107.6} & \textbf{129.9} & 59.6          & \textbf{81.4} & \textbf{104.4} & \textbf{128.5} & \textbf{144.3} & \textbf{171.9} & \textbf{86.7} & \textbf{133.0} & \textbf{174.3} & \textbf{218.1} & \textbf{255.2} & \textbf{304.1} & \textbf{823.9} & \textbf{1526.4} & \textbf{2152.0} & \textbf{2944.0} & \textbf{3560.2} & \textbf{4388.9} \\
    
\hline\hline

    \multicolumn{26}{c}{\textbf{CND}} \\
    \hline
    
    \multirow{2}{*}{\textbf{Method}}  & \multirow{2}{*}{\textbf{Mode}} & \multicolumn{6}{c|}{FF500} & \multicolumn{6}{c|}{ER500} & \multicolumn{6}{c|}{BA500} & \multicolumn{6}{c}{Yeast1} \\
    \cline{3-26}
    
    & & 40(4) & 80(8) & 120(12) & 160(16) & 200(20) & 240(24) & 40(4) & 80(8) & 120(12) & 160(16) & 200(20) & 240(24) & 40(4) & 80(8) & 120(12) & 160(16) & 200(20) & 240(24) & 20(2) & 40(4) & 60(6) & 80(8) & 100(10) & 120(12) \\
    \hline
    
    \multirow{3}{*}{TDE} & CPU & 3672.9 & 7201.5 & 10758.8 & 14508.8 & 18353.6 & 22275.7 
    & 3162.2 & 6201.3 & 9299.4 & 12771.5 & 15792.3 & 19319.5 
    & 3672.5 & 7111.4 & 10632.6 & 14533.0 & 18337.7 & 21938.5 
    & 29239.4 & / & / & / & / & / \\

    & S & \underline{816.3} & \underline{1625.7} & \underline{2433.1} & \underline{3233.6}    & \underline{4053.0}    & \underline{4913.9} 
    & \underline{738.3} & \underline{1457.3} & \underline{2197.7} & \underline{2923.7}    & \underline{3662.8}    & \underline{4400.8}  
    & \underline{758.4} & \underline{1505.8} & \underline{2269.6} & \underline{3027.2}    & \underline{3786.2}    & \underline{4573.6}  
    & \underline{5306.3} & \underline{10606.0} & \underline{15850.4} & \underline{21323.4} & \underline{26014.9} & \underline{31649.2}\\

    & M & \textbf{460.4} & \textbf{867.1} & \textbf{1279.6} & \textbf{1701.9} & \textbf{2133.3} & \textbf{2568.1}
    & \textbf{414.2} & \textbf{782.4} & \textbf{1154.8} & \textbf{1530.1} & \textbf{1911.9} & \textbf{2306.1}
    & \textbf{420.1} & \textbf{810.0} & \textbf{1198.3} & \textbf{1578.3} & \textbf{1984.5} & \textbf{2471.0}
    & \textbf{2934.8} & \textbf{5650.6} & \textbf{8865.3} & \textbf{10821.8} & \textbf{13317.7} & \textbf{15928.4}  \\

    \hline

    \multirow{3}{*}{CutOff}  & CPU & 832.5          & 2264.6         & 2461.8          & 3482.5          & \underline{4130.2}    & \underline{4993.3}    & 747.5          & 1994.5         & 2204.8          & 3091.4          & \underline{3640.4}    & \textbf{4394.8} & 661.8          & 1789.6         & 1963.8          & 2766.2          & \textbf{3237.9} & \textbf{3948.5} & 1766.0         & 3361.6         & 5156.5         & 6809.1          & 8933.5          & 9935.3          \\

    & S & \underline{420.3}    & \underline{992.4}    & \underline{1884.2}    & \underline{2725.7}    & 4139.3          & 5387.7          & \underline{375.9}    & \underline{932.1}    & \underline{1701.5}    & \underline{2700.0}    & 3807.1          & 4930.8       & \underline{400.4}    & \underline{1025.4}   & \underline{1797.4}    & \underline{2713.8}    & 3829.8          & 5320.3          & \underline{507.0}    & \underline{967.3}    & \underline{1496.4}   & \underline{2119.6}    & \underline{2619.5}    & \underline{3324.1}    \\

    & M & \textbf{321.9} & \textbf{814.2} & \textbf{1560.0} & \textbf{2436.7} & \textbf{3444.3} & \textbf{4836.5} & \textbf{305.5} & \textbf{782.3} & \textbf{1397.7} & \textbf{2262.0} & \textbf{3314.1} & \underline{4540.5} & \textbf{317.8} & \textbf{804.3} & \textbf{1494.1} & \textbf{2308.4} & \underline{3414.6}    & \underline{4926.7}    & \textbf{322.4} & \textbf{608.0} & \textbf{953.3} & \textbf{1391.8} & \textbf{1825.6} & \textbf{2231.1} \\

    \hline   
    \multirow{3}{*}{SixDST}  & CPU & 1469.9        & 2906.9         & 4232.5         & 5574.2         & 7168.8         & 8519.2         & 1051.2        & 2532.5         & 3667.5         & 4267.9         & 6240.6         & 6434.5         & 1416.2        & 2951.4         & 4240.8         & 5725.1         & 6965.9         & 8408.3         & /            & /            & /            & /            & /             & /             \\

    & S & \textbf{78.7} & \textbf{147.7} & \underline{216.5}    & \underline{289.5}    & \underline{351.3}    & \underline{430.9}    & \textbf{75.4} & \textbf{148.1} & \underline{222.8}    & \underline{296.7}    & \underline{366.7}    & \underline{432.8}    & \textbf{77.5} & \textbf{149.6} & \underline{222.6}    & \underline{295.5}    & \underline{366.0}    & \underline{422.3}    & \underline{367.0}    & \underline{732.0}    & \underline{1095.2}   & \underline{1454.6}   & \underline{1829.9}    & \underline{2188.0}    \\

    & M & \underline{142.0}   & \underline{183.2}    & \textbf{212.1} & \textbf{236.8} & \textbf{263.9} & \textbf{295.0} & \underline{140.4}   & \underline{182.7}    & \textbf{211.5} & \textbf{236.0} & \textbf{269.0} & \textbf{299.7} & \underline{140.4}   & \underline{181.1}    & \textbf{209.0} & \textbf{237.5} & \textbf{268.5} & \textbf{300.2} & \textbf{285.4} & \textbf{460.4} & \textbf{642.9} & \textbf{825.0} & \textbf{1014.9} & \textbf{1192.9} \\

\hline\hline

    \multicolumn{26}{c}{\textbf{NCA}} \\
    \hline
    
    \multirow{2}{*}{\textbf{Method}}  & \multirow{2}{*}{\textbf{Mode}} & \multicolumn{6}{c|}{CF900} & \multicolumn{6}{c|}{SF2000} & \multicolumn{6}{c|}{Cora} & \multicolumn{6}{c}{Citeseer} \\
    \cline{3-26}
    
    & & 20 & 40 & 60 & 80 & 100 & 120 & 20 & 40 & 60 & 80 & 100 & 120 & 20 & 40 & 60 & 80 & 100 & 120 & 20 & 40 & 60 & 80 & 100 & 120 \\
    \hline

    \multirow{3}{*}{GANI} & CPU & 1740.4 & 3088.3 & 4583.3 & 5735.3 & 7605.0 & 9231.0 & 9359.5 & 15745.8 & 23116.8 & 29678.1 & 35196.3 & 45482.3 & 7425.7 & 12282.6 & 17030.7 & 21957.5 & 27695.9 & 33137.9 & 6068.8 & 9047.5 & 12137.2 & 13468.6 & 17260.7 & 21248.0 \\

    & S & \textbf{498.9} & \underline{917.4}    & \underline{1110.3}    & \underline{1502.4}    & \underline{1727.0}    & \underline{2146.6}    & \textbf{943.4} & \textbf{1286.6} & \underline{1913.0}    & \underline{2289.6}    & \underline{2867.8}    & \underline{3341.3}    & \textbf{2168.7} & \textbf{3349.1} & \underline{4736.2}    & \underline{5619.7}    & \underline{6634.0}    & \underline{7991.5}    & \textbf{2835.1} & \textbf{3630.2} & \underline{4537.0}    & \underline{5218.2}    & \underline{5719.4}    & \underline{7769.8}    \\

    & M & \underline{675.4}    & \textbf{865.2} & \textbf{1056.4} & \textbf{1267.7} & \textbf{1392.7} & \textbf{1609.0} & \underline{1066.6}   & \underline{1428.8}    & \textbf{1611.0} & \textbf{1801.5} & \textbf{2006.6} & \textbf{2354.1} & \underline{2844.5}    & \underline{3493.5}    & \textbf{4126.5} & \textbf{4683.7} & \textbf{5301.8} & \textbf{5970.5} & \underline{3419.2}    & \underline{3970.4}    & \textbf{4402.6} & \textbf{4693.4} & \textbf{5126.0} & \textbf{6457.3} \\

    \hline

    \multirow{3}{*}{NCA-GA}  & CPU & 211.5         & 328.9          & 481.1          & 686.0          & 710.2          & 783.6          & 2617.1         & 4978.1         & 6643.0         & 9541.5         & 11279.5        & 11828.9         & 240.5         & 475.5          & 619.3          & 665.6          & 865.2          & 924.4          & 198.9         & 450.2          & 601.2          & 835.5          & 904.1          & 1015.1         \\

    & S & \textbf{89.3} & \underline{167.4}    & \underline{229.3}    & \underline{312.2}    & \underline{353.6}    & \underline{416.9}    & \underline{460.1}    & \underline{744.6}    & \underline{1030.6}   & \underline{1333.3}   & \underline{1628.0}   & \underline{1935.1}    & \textbf{71.0} & \underline{126.3}    & \underline{182.1}    & \underline{268.2}    & \underline{279.7}    & \underline{331.4}    & \textbf{61.5} & \textbf{105.0} & \underline{149.4}    & \underline{205.0}    & \underline{237.7}    & \underline{283.0}    \\

    & M & \underline{96.1}    & \textbf{116.7} & \textbf{155.1} & \textbf{178.2} & \textbf{216.6} & \textbf{267.3} & \textbf{277.4} & \textbf{437.8} & \textbf{591.9} & \textbf{716.2} & \textbf{889.4} & \textbf{1059.0} & \underline{96.5}    & \textbf{111.2} & \textbf{141.3} & \textbf{149.9} & \textbf{190.3} & \textbf{210.4} & \underline{85.9}    & \underline{109.6}    & \textbf{125.1} & \textbf{157.0} & \textbf{168.0} & \textbf{186.6} \\

\hline\hline

    \multicolumn{26}{c}{\textbf{LPA}} \\
    \hline
    
    \multirow{2}{*}{\textbf{Method}}  & \multirow{2}{*}{\textbf{Mode}} & \multicolumn{6}{c}{Dolphins}      & \multicolumn{6}{c|}{HD500}  & \multicolumn{6}{c|}{USAir97} & \multicolumn{6}{c}{Jazz}  \\
    \cline{3-26}
    
    & & 20 & 40 & 60 & 80 & 100 & 120 & 20 & 40 & 60 & 80 & 100 & 120 & 20 & 40 & 60 & 80 & 100 & 120 & 20 & 40 & 60 & 80 & 100 & 120 \\
    \hline

    \multirow{3}{*}{LPA-EDA} & CPU & \underline{45.8} & \underline{88.8} & \underline{128.2} & \underline{167.9} & \underline{174.5} & \underline{251.3} 
    & \textbf{462.6} & \textbf{904.3} & \underline{1292.6} & \underline{1746.4} & \underline{2185.0} & \underline{2610.7} 
    & 3367.2 & 6791.7 & 9621.6 & 12835.2 & 14104.0 & 18476.0 
    & 2432.8 & 4497.1 & 6804.3 & 8973.1 & 10002.5 & 13434.2 
    \\
    
    & S & 56.7 & 106.4 & 151.0 & 211.0 & 263.3 & 310.8 
    & 1084.1 & 1917.0 & 2754.6 & 3641.3 & 4322.8 & 4743.1     
    & \underline{1336.8} & \underline{2599.0} & \underline{3754.2} & \underline{4658.5} & \underline{6157.1} & \underline{7504.4} 
    & \underline{1196.8} & \underline{2580.8} & \underline{3878.1} & \underline{5196.7} & \underline{7309.2} & \underline{9133.3}
    \\
 
    & M & \textbf{45.3} & \textbf{72.2} & \textbf{101.8} & \textbf{114.8} & \textbf{149.8} & \textbf{168.7} 
    & \underline{564.8} & \underline{920.4} & \textbf{1245.2} & \textbf{1561.6} & \textbf{1905.5} & \textbf{2212.3}  
    & \textbf{717.9} & \textbf{1225.8} & \textbf{1881.5} & \textbf{2458.1} & \textbf{3045.7} & \textbf{3654.3}
    & \textbf{566.8} & \textbf{1263.1} & \textbf{2213.5} & \textbf{2816.1} & \textbf{3503.8} & \textbf{4382.6}
    \\

    \hline
    \multirow{3}{*}{LPA-GA}  & CPU & \textbf{22.9} & \textbf{34.9} & \textbf{44.4} & \textbf{52.2} & \textbf{59.5} & \textbf{62.1} 
    & \textbf{263.0} & \textbf{395.8} & \textbf{486.7} & \textbf{568.2} & \textbf{635.2} & \textbf{693.6}  
    & 1181.7 & 1794.2 & 2229.8 & 2627.3 & \underline{2917.7}    & \underline{3128.9} 
    & 846.3 & 1292.8 & 1662.8 & 1897.4 & \underline{2112.4} & \underline{2278.3}
    \\

    & S & \underline{32.8} & 57.9 & 85.9 & 112.7 & 139.0 & 170.6 
    & 744.9 & 1249.0 & 1838.3 & 2187.7 & 2752.4 & 3156.4 
    & \underline{720.5} & \underline{1329.7} & \underline{2010.8} & \underline{2696.2} & 3301.3 & 4119.6 
    & \underline{462.2} & \underline{893.6} & \underline{1304.1} & \underline{1681.5} & 2213.2 & 2466.3 
    \\

    & M & 34.2 & \underline{47.0} & \underline{61.1} & \underline{76.1} & \underline{92.2} & \underline{102.3} 
    & \underline{433.7} & \underline{607.7} & \underline{797.4} & \underline{991.1} & \underline{1172.2} & \underline{1468.7}  
    & \textbf{431.0} & \textbf{705.9} & \textbf{1082.9} & \textbf{1326.0} & \textbf{1658.9} & \textbf{2068.5} 
    & \textbf{266.5} & \textbf{493.8} & \textbf{705.0} & \textbf{873.7} & \textbf{1097.5} & \textbf{1309.8}
    \\

\hline\hline
\end{tabular}
}
\end{table*}

\subsection{Performance On the Different Acceleration Mode}
\label{PEAM}
Here, we evaluate the acceleration performance of GAPA on GPU when processing different tasks and datasets with different scales. Specifically, this experiment is conducted with 4 acceleration modes, 10 GA-based algorithms on 4 PSSO tasks, and 12 datasets.~\cref{table:Whole Table} contains the primary results, showing the processing efficiency of each algorithm and two performance evaluation results under the corresponding tasks. Meanwhile, the gray-shaded areas highlight the acceleration efficiency results of the algorithm. A smaller time value indicates better acceleration performance. The optimal computing times are marked in bold, while the suboptimal computing times are underlined. Q, NMI, PC(G)\cite{cutoff}, Accuracy, ASR(Attack Successful Rate), Precision, and AUC represent the performance indicators for the CDA, CND, NCA, and LPA tasks, respectively. Additionally, for experimental results with a running time exceeding 12 hours, we mark them with \textbf{/} and do not perform statistics.

From the whole perspective, \cref{table:Whole Table} reveals that the performance of all optimization algorithms on relevant tasks is basically consistent and produces significant acceleration, especially as the data size grows. From the perspective of the dataset, initial experiments at small scales of datasets may favor the CPU, but GPU acceleration rapidly overtakes it as the scale increases, achieving twofold to tenfold speedup, especially in M mode. From the perspective of different acceleration modes, a notable observation is that M is usually the optimal acceleration mode, while MNM mode and S mode will become sub-optimal acceleration modes as the data size increases. This situation shows that heavyweight computations can better utilize acceleration resources simultaneously, ignoring the data transfer costs. From the perspective of different tasks, GAPA has demonstrated significant acceleration effects on the CDA, CND, and NCA tasks. However, since the algorithm used for the LPA task has already been fully optimized, the acceleration effect observed in this task is moderate. Next, we conduct specific analyses on different tasks.

In the CDA task, the CDA-EDA algorithm achieved a significant acceleration of nearly 10x, while CGN and QAttack achieved up to 4x acceleration, respectively. It is worth noting that the NMI values of CDA-EDA and QAttack show large fluctuations, primarily because these two algorithms are not specifically optimized for NMI. In the CND task, SixDST outperforms the TDE and CutOff algorithms on small-scale data sets, with speedups of approximately 6x and 4x, respectively. On large-scale data sets, SixDST is approximately 10x faster than TDE and 16x faster than CutOff because the computational time for large-scale matrix operations increases significantly as the data size grows. SixDST only relies on matrix operations, so its computational cost on the Yeast1 dataset is close to that of CutOff. For the NCA task, the GANI and NCA-GA algorithms achieved nearly 17x and 13x acceleration, respectively. NCA-GA uses perturbed network edges, leading to a significant increase in processing time on the SF2000 network, which has numerous edges. In contrast, GANI involves additional computational steps beyond using GA to generate edges connecting fake nodes, which consume considerable time. We significantly reduced the computational time of GANI on SF2000 by optimizing redundant edge calculations, but the effectiveness of this optimization is influenced by the network node scale. Additionally, in SM mode, the data transmission overhead of GANI is negligible compared to the computational cost, achieving an ideal acceleration effect. For the LPA task, although LPA-EDA has fully optimized the algorithm in its original form, we still achieved an acceleration of up to about 5x, and LPA-GA achieved an acceleration of about 3x. It is worth mentioning that when the data processing cost becomes high enough that the overhead caused by data transmission and process creation and destruction can be ignored, the MNM mode produces an acceleration effect, as verified in experiments on the Jazz dataset.

Overall, GAPA has demonstrated significant acceleration for the GA-based PSSO problem while maintaining the stable performance of the original algorithm. However, M mode, which provides the best acceleration, is limited in fully utilizing acceleration resources due to the inherent constraints of the algorithm. Moving forward, we are supposed to optimize GAPA further to monitor acceleration resource usage and adaptively adjust the number of processes, maximizing both resource utilization and acceleration efficiency.

\subsection{The Analysis of Population Size}
\label{PTIP}

Here, to explore the impact of population size on GPU acceleration, we conduct extensive experiments on several population size settings with original and optimized algorithms across 4 tasks. Considering that the SM and MNM modes are extensions of the S and M modes, and the performance of them in acceleration experiments has not been particularly outstanding, we only conduct experiments on S and M modes. \cref{table:Scale} shows the results on 12 datasets and 10 algorithms across 4 tasks. Specifically, in the CND task, due to the time-intensive computation of the TDE algorithm, we restrict its population size to [2, 4, 6, 8, 10, 12] in Yeast1 and [4, 8, 12, 16, 20, 24] in other datasets.

In general, the calculation time of the original algorithm demonstrates a nearly linear relationship with population size, and the computation time of the optimized algorithms follows a similar linear trend. Moreover, as the dataset scales grow, the acceleration effect of all optimization algorithms becomes significantly more pronounced. However, it is important to note that the gap between the performance of all optimization algorithms and the original algorithm is minimal when the population size is small. In some cases, certain optimization algorithms even perform slower than the original algorithm when the population size is small. As the population size increases, the processing time of most optimization algorithms grows at a slower rate compared to the original algorithm. Consequently, with larger population sizes and datasets, the acceleration effect of GAPA becomes increasingly significant. Furthermore, the performance of these algorithms across different datasets is generally consistent with the results observed in the acceleration performance experiments. Next, we provide a more detailed analysis of the different tasks.

In the CDA task, we observe that when the population size of QAttack is below 100, the processing speed of the original algorithm is superior to that of S mode and M mode. Furthermore, when the population size is below 40, S mode outperforms M mode. This indicates that, for lightweight computations, GPU acceleration can actually become a burden, leading to inefficiencies. A similar phenomenon is observed in CGN, where the burden on the GPU increases as the dataset size scales up. In the CND task, we note that when the genetic size is below 200, CutOff shows a noticeable acceleration in both S mode and M mode, with an almost 3x speedup observed when the genetic size is 80. However, once the population size exceeds 200, the processing speed of the original algorithm becomes faster than the optimized algorithm. This behavior can be attributed to the inefficiencies of large-scale population transfer, which is much slower than directly processing the data. On the Yeast1, the data transmission costs are negligible compared to the total computational costs, and thus the acceleration benefits of the optimization algorithm become more significant under larger population sizes. The NCA and LPA tasks exhibit similar characteristics to the CDA and CND tasks, where the effect of optimization algorithms is more pronounced as the population size increases.

In summary, the acceleration effect of GAPA generally increases as the population size grows, particularly for larger-scale datasets. This experiment also confirms that CPU outperforms GPU in lightweight computations while highlighting the significant advantages of GAPA in handling heavyweight computations that fully utilize the GPU acceleration resources.

\begin{figure}[!t]
\centering

\includegraphics[width=0.45\textwidth]{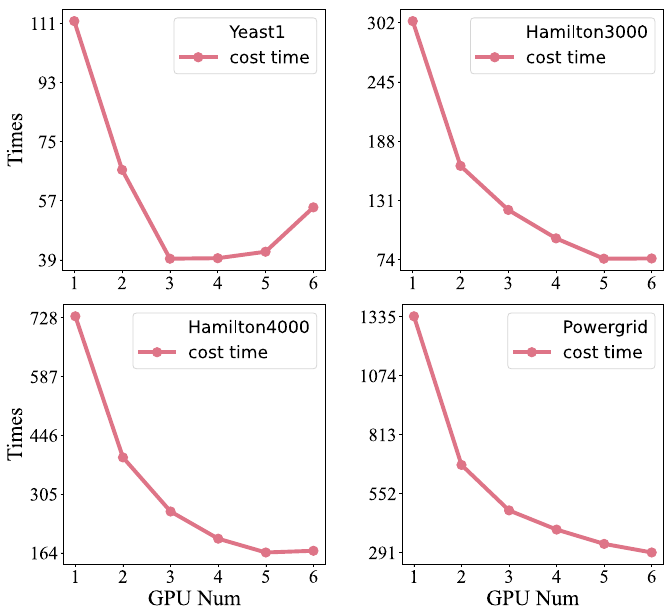}

\caption{The diagram illustrates the computational acceleration of the SixDST across four datasets, showing the performance improvement with varying numbers of GPUs, from $1-6$.}
\label{fig:MutiGPU}
\end{figure}

\subsection{The Analysis of Distributed Acceleration}
\label{MultiGPU}
In order to evaluate the acceleration effect of GAPA on GA-based PSSO problems in a distributed environment, we conducted distributed acceleration experiments using different numbers of GPUs on the Yeast1, Hamilton3000, Hamilton4000, and Powergrid, applying both S and M modes, respectively. For SixDST, the iteration number was set to 200, and the $pop\_size$ was set to 60. The experimental results, presented in Figure \cref{fig:MutiGPU}, show the effect of varying GPU numbers on computation time, displayed as a line chart. From the overall trend, it is clear that as the number of GPUs increases, the computation time for SixDST decreases significantly. It is easy to conclude that 1) GAPA can significantly enhance the acceleration effect of the algorithm in a distributed environment; 2) The larger the dataset, the more pronounced the acceleration effect of GAPA. Particularly, in Yeast1, when the number of GPUs is less than 3, the computing time decreases noticeably with each additional GPU, showing a clear linear acceleration. However, once the number of GPUs reaches between 3 and 4, the computing time stabilizes, with little change. Beyond 4 GPUs, the computing time continues to decrease, but the rate of acceleration slows down. This phenomenon aligns with our expectations based on the acceleration performance and population size. As more GPUs are added, the data transmission cost increases significantly. When 6 GPUs are used, the data transmission cost surpasses the computation cost, leading to a slight increase in overall computing time. Similar trends are observed with Hamilton3000 and Hamilton4000. For both datasets, the acceleration effect saturates when 5 GPUs are used, and further increases in GPU number do not bring substantial improvements. Moreover, when the data size increases, particularly on the Powergrid dataset, the calculation time curve shows a sublinear relationship with the number of GPUs. This suggests that after reaching a certain threshold, adding more GPUs results in diminishing returns, particularly when the data size is large and the GPU processing power becomes a limiting factor.

Above all, it is evident that deploying more GPUs can still provide acceleration benefits as the number of nodes in the network increases. This insight can guide the optimization of resource allocation, ensuring efficient utilization of acceleration resources and achieving an optimal balance between computational efficiency and data transmission overhead.


\begin{figure}[!t]
\centering
\includegraphics[width=0.45\textwidth]{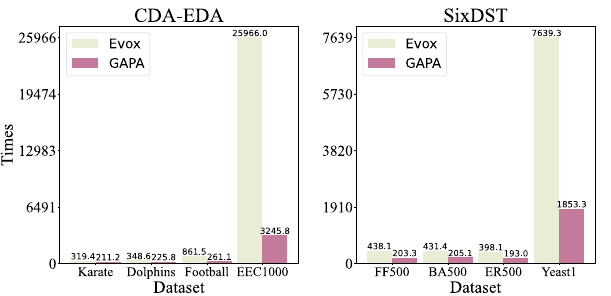}
\caption{The figure illustrates the acceleration effects of CDA-EDA and SixDST, showing that applying GAPA across 4 different datasets provides significant advantages over using Evox.}
\label{fig:Cons}
\end{figure}

\subsection{Comparison with Evox}
\label{Comparative}
To further demonstrate the superiority of GAPA, we compared it with Evox, a state-of-the-art evolutionary computing acceleration library built on JAX. In this experiment, we utilized the general computing framework provided by Evox. To highlight the comparison, we selected two computation-intensive algorithms (CDA-EDA and SixDST) with matrix calculation, which can fully utilize acceleration resources.

\cref{fig:Cons} shows the comparison of acceleration effects of CDA-EDA and SixDST under two acceleration frameworks across eight datasets. Due to the considerable computing time when no acceleration framework is used, which causes the histogram trend to be unclear, we list the algorithm running times without acceleration as follows: NCA-EDA [248.6, 815.7, 2159.3, /], SixDST [4232.5, 4240.8, 3667.4, /]. It is clear that both GAPA and Evox achieve significant acceleration effects, where GAPA and Evox reach 6$\times$, 3$\times$ acceleration on NCA-EDA, and 20$\times$, 10$\times$ acceleration on SixDST, respectively. The overall acceleration effect of GAPA is twice of Evox. In particular, GAPA demonstrates the most significant acceleration on the EEC1000 and Yeast1 datasets, achieving speedups of 8$\times$ and 4$\times$ compared to Evox, respectively. Based on these experimental results, It is easy to conclude that 1) As the data scale increases, GAPA achieves more significant acceleration effects; 2) The GAPA acceleration framework outperforms Evox for GA-based PSSO problems, making it more suitable for the acceleration needs of large-scale datasets.

\section{Conclusion}
\label{Conclusion}
In this paper, we introduce and implement GAPA, the first GA acceleration framework tailored for the PSSO problem. GAPA utilizes parallel acceleration strategies to reduce computational time in GA-based PSSO problems significantly. Additionally, it provides a specific design example for fitness calculation functions, serving as a reference for users while highlighting the advantages of parallel acceleration. Finally, we conduct acceleration optimization and experimental analysis on ten algorithms across four PSSO subtasks, with results demonstrating that the proposed strategies greatly enhance the execution efficiency of GA. 

Looking ahead, as the field of perturbed substructure optimization continues to evolve, GAPA aims to optimize additional GA algorithms within this domain and make them widely accessible. In future developments, we aim to simplify the acceleration framework for easier adoption further while dynamically identifying and managing acceleration resources to maximize their utilization.

\nocite{*}
\bibliography{ref.bib}
\bibliographystyle{IEEEtran}

\end{document}